# Ellogon: A New Text Engineering Platform

**Georgios Petasis, Vangelis Karkaletsis, Georgios Paliouras,
Ion Androutsopoulos and Constantine D. Spyropoulos**

Software and Knowledge Engineering Laboratory,
Institute of Informatics and Telecommunications,
National Centre for Scientific Research (N.C.S.R.) "Demokritos",
P.O. BOX 60228, Aghia Paraskevi,
GR-153 10, Athens, Greece.
e-mail: {petasis, vangelis, paliourg, ionandr, costass}@iit.demokritos.gr

**Abstract**

This paper presents *Ellogon*, a multi-lingual, cross-platform, general-purpose text engineering environment. *Ellogon* was designed in order to aid both researchers in natural language processing, as well as companies that produce language engineering systems for the end-user. *Ellogon* provides a powerful TIPSTER-based infrastructure for managing, storing and exchanging textual data, embedding and managing text processing components as well as visualising textual data and their associated linguistic information. Among its key features are full Unicode support, an extensive multi-lingual graphical user interface, its modular architecture and the reduced hardware requirements.

## 1. Introduction

In this paper we describe *Ellogon*, a new text engineering platform developed by Software and Knowledge Engineering Laboratory of the Institute of Informatics and Telecommunications, N.C.S.R. "Demokritos", Greece. *Ellogon* is a multi-lingual, cross-platform, general-purpose text engineering environment, developed in order to aid both researchers who are doing research in the natural language field or computational linguistics, as well as companies that produce and deliver language engineering systems.

During the last decade, a large number of software infrastructures aiming at facilitating R&D in the field of natural language processing have been presented. According to the model used to associate the textual data with the corresponding linguistic information, these infrastructures can be classified into the following three types (Cunningham, 1997):

- **Additive or Mark-up based:** the linguistic information is added to the text using a mark-up tagging scheme. A known example of this category is the LT-NSL toolkit from the University of Edinburgh (Thompson and McKelvie, 1996) (McKelvie *et. al.* 1997).
- **Referential or Annotation based:** the linguistic information is stored separately from the textual data, having references back to the original texts. GATE (Cunningham, 1997), the TIPSTER project (Grishman, 1996) as well as our platform are examples of this category.
- **Abstraction based:** the textual data is preserved only as parts of an integrated data structure that represents information about the text in a uniform theoretically motivated model. A representative of this category is the ALEP system (Simkins, 1994).

A fourth category may be added, in order to include systems that provide only communication and control infrastructure without imposing a uniform representation scheme among components, such as the TalLab platform (Wolinski *et al.* 1998) or the ICE architecture (Amtrup, 1995).

The rest of the paper is organised as follows: In section 2 some of the existing text engineering platforms are briefly described. In section 3 *Ellogon* is presented. Finally, in section 4 some concluding remarks are presented.

## 2. Related Work

The LT-NSL and LT-XML tools are examples of the additive approach. They are based on SGML and XML respectively. These tools provide a specialised application programming interface (API) with the help of which programs can request, add or modify linguistic information in documents encoded in SGML/XML. The components that actually perform the linguistic processing are SGML-aware executables (using the provided SGML API) that communicate with each other with the use of pipes. One of the most important advantages of this approach is that each linguistic component can load from the SGML document only the information that requires or can recognise, resulting in a system with small memory requirements. On the other hand, the need for re-parsing the SGML document by each module may increase the processing time. In addition, there may be problems related to SGML. For example, although tree-structured information can be easily encoded, graph-structured information is difficult to represent.

GATE is probably the most widely used text engineering platform. GATE follows the referential or annotation-based approach, based on the TIPSTER data model. Under GATE, linguistic information is stored independently of the textual data, in a database (GDM – the GATE Document Manager). GATE also offers a specialised object oriented API for retrieving, adding or modifying the associated linguistic information. Besides this core functionality, GATE presents some additional features such as the ability to integrate linguistic processing components at run-time, a graphical user interface, a set of visualisation tools and a set of some generic tools like comparison tools or SGML import/output filters. On the other hand, GATE presents large requirements of operating system resources and lacks support for non-Latin languages[1] (Cunningham, 2000).

The Advanced Language Engineering Platform (ALEP) provides infrastructure for accessing linguistic processing and text handling tools, resources and applications. ALEP utilises a neutral feature-based, unification enabled formalism for storing all needed linguistic information. ALEP also offers the ability to integrate external linguistic processing components and provides many tools for converting the utilised formalism into PROLOG as well as for debugging the embedded components. Finally, a graphical user interface is offered, based on EMACS (a Unix LISP-based text editor) and Motif (a Unix graphical toolkit). On the other hand, ALEP imposes a specific formalism and is dependent on a specific operating system (Unix). While ALEP provides an open framework where new formalisms can be embedded, the provided formalism can be only used for developing particular types of resources (like grammars or lexicons) and for performing a particular set of linguistic tasks (or performing them in a particular way).

The TalLab platform does not impose a single standard for representing linguistic information and as a result it does not restrict the way a component is implemented. It is based on a multi-agent approach, reusing the operating system wherever possible. Each component is implemented as an agent that can communicate with other agents in an asynchronous mode. One of the most important advantages of this approach is the fact that it can operate in mission-critical environments. Possible failure of an agent does not halt the entire system: the problem cause can be eliminated (like rejecting a problematic document) and the agent can be restarted. On the other hand, the lack of a communication standard makes difficult the integration of new components, as each component has to communicate directly with existing agents that possibly utilise various communication protocols.

The main reason that led to the development of *Ellogon* was the inadequacy of existing platforms to support some important properties in text engineering. These features include the ability to support a wide range of languages through Unicode, to function under all major operating systems, to have as few hardware requirements as possible, to be based on a modular architecture that enables parts to be embedded in other systems and to provide an extensible, easy to use and powerful user interface. *Ellogon* supports many of these requirements to a large extent.

## 3. Ellogon

*Ellogon* belongs to the category of referential or annotation based platforms. Based on the TIPSTER data model, *Ellogon* provides infrastructure for:
- Managing, storing and exchanging textual data as well as the associated linguistic information.
- Creating, embedding and managing linguistic processing components.
- Facilitating communication among different linguistic components by defining a suitable API.
- Visualising textual data and associated linguistic information.

The architecture of *Ellogon*, the utilised data model and the linguistic processing components as well as some key features of *Ellogon* are presented in the following subsections.

### 3.1. Ellogon Architecture

*Ellogon* consists of mainly three subsystems (Figure 1):
- A highly efficient core – the Collection and Document Manager (CDM), developed in C++, which implements all the basic functionality. Its design is based on the TIPSTER architecture[2] and its main responsibility is to manage the storage of the textual data and the associated linguistic information and to provide a well-defined programming interface that can be used in order to retrieve/modify the stored information. This core can be easily embedded in other applications.
- A powerful and easy to use graphical user interface (GUI). This interface can be easily tailored to the needs of the end user.
- A modular plugable component system. All linguistic processing within the platform is performed with the help of external, loaded at run-time, components.

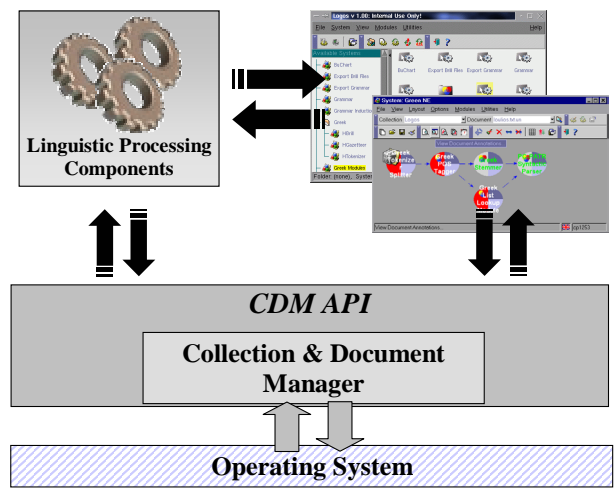

Figure 1: *Ellogon* Architecture.

### 3.2. Ellogon Data Model

*Ellogon* shares the same data model as the TIPSTER architecture. The central element for storing data in *Ellogon* is the *Collection*. A collection is a finite set of *Documents*. An *Ellogon* document consists of *textual data* as well as *linguistic information about the textual data*. This linguistic information is stored in the form of *attributes* and *annotations*.

An attribute associates a specific type of information with a typed value. For example, the part-of-speech of a word can be represented with the following attribute:

pos = (STRING) "noun"

---

[1] However, GATE version 2 will provide complete support for non-Latin languages, as it will offer complete Unicode support.

[2] Although *Ellogon* is based on the TIPSTER architecture, it is not strictly TIPSTER compliant. For example, currently the API does not provide the object-oriented framework defined by the TIPSTER architecture.

As can be seen from this example, attribute values are typed. Currently supported values are strings, sets of strings and annotations.

An annotation associates arbitrary information (in the form of attributes) with portions of textual data. Each such portion, named *span*, consists of two byte offsets denoting the start and the end characters of the portion, as measured from the first character of some textual data. Annotations typically consist of four elements (see Figure 2):

- *A numeric identifier*. This identifier is unique for every annotation within a document and can be used to unambiguously identify the annotation.
- *A type*. Annotation types are textual values that are used to classify annotations into categories.
- *A set of spans* that denote the range of the annotated textual data.
- *A set of attributes*. These attributes usually contain the necessary linguistic information.

| This is a simple sentence. |
| :---: |
| 0....5....10...15...20...25 |

| | | Annotations | | | |
| :---: | :---: | :---: | :---: | :---: | :---: |
| | | Span Set | | | |
| ID | Type | Span 1 | | Span 2 ... | Attributes |
| | | Start | End | Start | End | |
| 0 | token | 0 | 4 | | | type=EFW, pos=PN |
| 1 | token | 5 | 7 | | | type=ELW, pos=VB |
| 2 | token | 8 | 9 | | | type=ELW, pos=IDT |
| 3 | token | 10 | 16 | | | type=ELW, pos=ADJ |
| 4 | token | 17 | 25 | | | type=ELW, pos=NN |
| 5 | token | 25 | 26 | | | type=PUNC, pos=. |
| 6 | sentence | 0 | 26 | | | constituents=[0 1 2 3 4 5] |
| 7 | link | 0 | 4 | 17 | 25 | constituents=[0 4] |

Figure 2: Example of a sentence and its relevant annotations.

### 3.3. Ellogon Components

For most users of *Ellogon*, the central point of interest is the linguistic processing that can be carried out within it. *Ellogon* provides a generic framework where external components can be embedded. These components are organised into *Systems* for performing some specific task. The tasks can range from basic linguistic tasks, such as part-of-speech tagging or parsing, to application level tasks, such as information extraction or machine translation.

A component consists mainly of two parts. The first part is responsible for performing the desired linguistic processing while the main responsibility of the second component part is to interface the linguistic processing sub-component with *Ellogon*, through the provided API. Components can appear either as *wrappers* or as *native components*. Wrappers usually provide the needed code in order to interface an existing independent implementation of a linguistic processing tool to the *Ellogon* platform. Native components on the other hand are processing tools specifically designed for use within the *Ellogon* platform. Usually, in such components the two component parts cannot be easily identified or separated.

Each component is associated with a set of pre-conditions and a set of post-conditions. Pre-conditions declare the linguistic information that must be present in a document before this specific component can be applied to it. Post-conditions describe the linguistic information that will be added in the document as a side effect of processing the document with this specific component. *Ellogon* uses these two sets in order to establish relations among the various components.

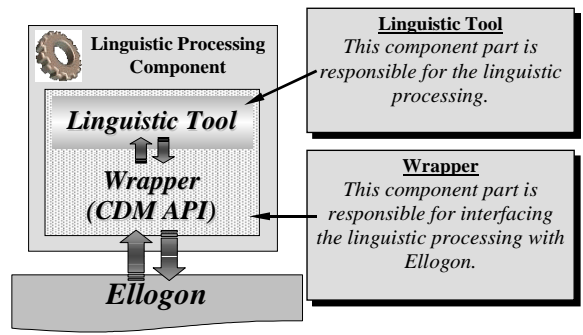

Figure 3: Component Structure.

Each component can also specify a set of parameters, as well as a set of viewers. Parameters represent various run-time dependent options (such as the location of a file containing the grammar of a syntactic parser). They can be edited by the end user through the graphical interface and are given to the component every time it is executed. A component can also specify a set of predefined viewers, in order to present in a graphical way the linguistic information produced during the component execution. Examples of available viewers are shown in figures 6 – 9.

Creating components can be easily done through the *Ellogon* GUI. Currently, *Ellogon* components can be developed in two languages, C++ and Tcl. The *Ellogon* GUI offers a specialised dialog where the user can specify various parameters of the component he/she intends to create, including its pre/post-conditions. Then *Ellogon* creates the *skeleton* of the new component that will handle all the interaction with the *Ellogon* platform. If the language of the component is C++, a *Makefile* for compiling the component under Unix will also be created. Besides creating a skeleton, *Ellogon* tries to facilitate the development of the component by allowing the developer to edit the source code and re-load the specific component into *Ellogon* from its GUI.

### 3.4. Ellogon key features

In the following paragraphs, we briefly present some of the most important aspects of the *Ellogon* text engineering platform.

#### 3.4.1. Support of multiple languages

The fact that *Ellogon* offers full Unicode support (in both its core unit CDM as well as in its GUI) provides the ability to properly support a wide range of languages. *Ellogon* includes a large number of input/output filters for various encodings, such as the ISO-8859-* encodings or the encodings used under Microsoft Windows or Apple Macintosh. Additionally, components can be classified according to the language they support and can utilise the utilities provided by the API in order to convert textual data among various encodings. Finally, *Ellogon* provides an internationalised GUI[3] that has been designed to facilitate the integration of additional languages, even by the end user.

### 3.4.2. Portability

Supporting all the major operating systems has always been a shortcoming of many of the existing text engineering platforms. *Ellogon* on the other hand, offers native ports to many operating systems and has been extensively used and tested under Unix (Solaris 2.6 & 7, Red Hat Linux 6 & 7) and Microsoft Windows (95, 98, Me, NT 4.0, NT 2000 & XP). Additionally, *Ellogon* aims to provide a unified view of various operating system specific tasks under all supported operating systems. For example, pipelines and file redirections are emulated under Microsoft Windows or filenames can be specified using the Unix notation under all supported operating systems. Finally, the provided graphical interface provides exactly the same functionality under the various supported operating systems.

### 3.4.3. Advanced GUI

*Ellogon* offers an extensive and powerful multi-lingual user interface. This GUI provides users with the ability to manage Collections/Documents/Systems, to visualise linguistic information with an extensible set of visualisation tools, to develop and integrate linguistic components, to browse documentation and of course, to do linguistic processing of textual data using various modes. Finally, the user interface can be adapted to meet specific needs, such as systems dedicated to specific linguistic processing tasks.

### 3.4.4. Modular Architecture

*Ellogon* is based on a modular architecture that allows the reuse of *Ellogon* sub-systems in order to ease the creation of applications targeting specific linguistic needs.

*Ellogon's* core component – CDM – is implemented as a separate library that can be dynamically loaded if the underlying operating system offers such ability. This library can be independently embedded inside any application that can call functions from libraries, following the C++ naming conventions. Examples of embedding CDM under various applications include Microsoft Word[4], Tcl and Java.

Actually, the whole *Ellogon* platform is based on the idea of extending the Tcl scripting language by embedding CDM into it. The graphical interface of *Ellogon*, is based on the cross-platform graphical toolkit Tk, included in Tcl. Developed in a scripting language it can be easily extended (even components can utilise/extend it) or modified to create specialised applications.

The choice of incorporating Tcl into *Ellogon* has played an important role in its development. *Ellogon* takes advantage of many features from Tcl, like Unicode support, data structures and its cross-platform abstraction layer. Other cross-platform technologies (Java included) were not found adequate for all the requirements a platform like *Ellogon* must have. Initial work in embedding CDM into Java[5] has been done, but results were not encouraging. Memory usage increased significantly, while processing time was not greatly improved compared to Tcl. In addition, utilising Java graphical capabilities was found slower and far more complex than utilising the corresponding capabilities offered by Tcl. Additional problems were also discovered, like the incompatibilities among various Java implementations from different vendors and the difficulty to design code that cooperates well with all of them[6]. Currently, Tcl appears to offer the best cross-platform support, although we expect that Java may also be improved in future versions.

### 3.4.5. Interoperability with other platforms

*Ellogon* facilitates the reuse of linguistic processing tools. Under this framework, *Ellogon* allows the reuse of existing linguistic components, even if they currently operate under other platforms. Of course, this is not an easy task, as it usually requires re-implementing the platform that is emulated. Currently, *Ellogon* offers only one compatibility layer, enabling the incorporation and embedding of GATE (version 1.x) components (CREOLE objects) written in the Tcl language.

Its development was mainly driven by the fact that we have been users of the GATE platform for many years and almost all of our components were working under GATE. This compatibility layer gave us the ability to cooperate with other groups that use GATE, facilitating the development of components that operate under both platforms and the ability of exchanging textual data and the associated information. This compatibility layer offers the ability to execute GATE Tcl components, to import/manage/export GATE collections and to import/manage/export GATE systems. Currently, no support exists for GATE modules developed in C++.

### 3.4.6. Memory compression

The use of memory by a text-engineering platform is a very important aspect, as it usually determines the size of textual data that can be processed under this platform. Under *Ellogon*, this requirement is far more important, as the use of Unicode can increase memory requirements by simply changing from a language that requires fewer bytes per character (like English) into a language needing more bytes per character (like Greek). *Ellogon* tries to decrease its memory requirements by incorporating a memory compression scheme. Initial measurements have shown that *Ellogon* uses less memory for performing the same tasks than other (TIPSTER-based) platforms.

### 3.4.7. Execution server

Executing other programs from an application is not always an easy task, as it is usually associated with a large number of problems. Except of utilising platform specific features (like pipelines or file redirections), usually a large amount of memory is needed (especially under Unix) in order for a program to execute another program. For all these reasons, *Ellogon* incorporates an execution server: when *Ellogon* initialises, it also initialises a separate process that specialises in executing external programs. When the need to execute an external program arises, the operating system needs to replicate the execution server instead of the main *Ellogon* process that may request hundreds of megabytes of memory.

---

[3] Currently, the provided GUI languages include only English and Greek.
[4] In order to embed CDM under Microsoft Word we utilise the Active-X technology, with CDM exported as an Active-X component.
[5] Sun JDK 1.3.0 and IBM JDK have been used. CDM was embedded using the Java Native Interface (JNI).
[6] Currently, only Microsoft Windows, Sun Solaris and Linux are supported.

### 3.4.8. HTTP Server

Finally, an interesting feature of *Ellogon* is the ability to act as a *Web server*, offering the ability to expose the linguistic functionality implemented by its components through the *HTTP* protocol. Thus, other instances of the *Ellogon* platform, possibly running on remote computers, or even single users through a web browser, can upload textual data to an *Ellogon* server, use its components to process the data and download the results back to the remote computer as a Web page.

### 3.5. Systems that use Ellogon

*Ellogon* has been constantly used by SKEL laboratory in research projects (four of which are presented below). *Ellogon* will be used soon by other research organisations (project partners of SKEL), in order to be better evaluated. We hope to collect valuable information about its use that will enable us to further improve *Ellogon* along many directions.

### 3.5.1. GIE

Greek Information Extraction (GIE) was a bilateral project between our laboratory and the University of Sheffield (Karkaletsis *et. al.* 1998) aiming at the creation of a Greek information extraction system based on the English VIE system distributed with GATE. The resulting system was also embedded in the GATE platform. This system gave us the ability to test and evaluate the GATE compatibility mode of *Ellogon*, since the GIE system also runs unmodified under *Ellogon*.

### 3.5.2. MITOS

MITOS is an R&D project[7] that combines techniques from information filtering to classify incoming news articles, as well as techniques from information extraction to extract factual information from financial news articles, which is then stored into a database (e.g. buyer, company bought). *Ellogon* was used as the development platform for the linguistic processing and information extraction components. It was also used to develop user-friendly applications for information extraction and for annotating training data. These applications are currently used successfully by users with no linguistic or NLP background.

### 3.5.3. SCHEMATOPOIESIS

In the context of the Greek R&D project SCHEMATOPOIESIS, *Ellogon* was used to develop the first prototype controlled language checker for Greek in order to assist Greek technical writers as well as to facilitate translation from Greek to other languages[8]. The project covered technical documents from the domain of computer equipment. *Ellogon* was used not only as the development platform for the checker, but also as *a mean for embedding the checker under Microsoft Word*, allowing the user to check his/her documents in a similar way as a spell/syntax checker. The architecture of the controlled language checker (Petasis *et al.* 2001) is shown in Figure 4.

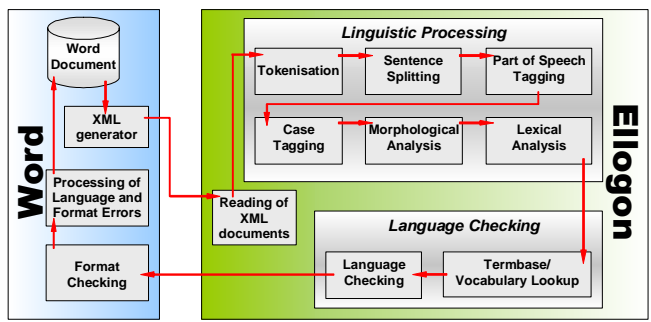

Figure 4: The architecture of the SCHEMATOPOIESIS controlled language checker. *Ellogon* was used as a "vehicle" for embedding the checker under Word. Communication between Word and *Ellogon* was achieved by means of XML and DDE messages.

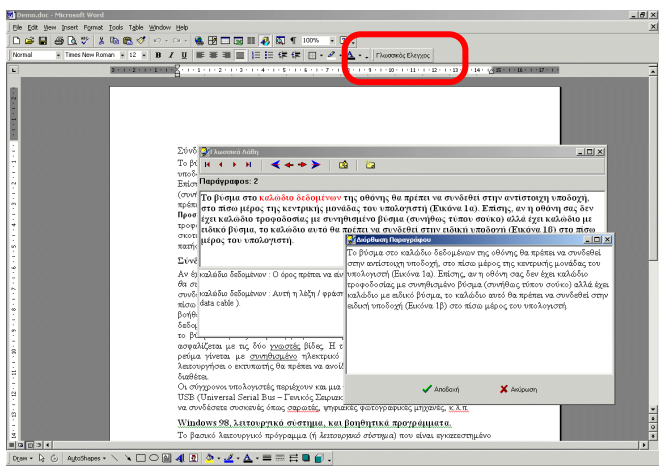

Figure 5: The interface for correcting errors identified by the SCHEMATOPOIESIS checker. *Ellogon* is completely hidden to the end user. The user simply presses a toolbar button to process the document with *Ellogon* and see/correct the identified errors within Word.

All the components related to linguistic processing and language checking were running under *Ellogon*, where as the components for generating an XML-based representation of the word document, the components that performed the formatting checks and the components that marked the identified errors on the document were running under MS Word. The communication between MS Word and *Ellogon* was achieved with the use of either ActiveX or DDE, services supported by both MS Word and the Windows version of *Ellogon*. The reason we have evaluated two communication methods is related to ro-bustness. Under the ActiveX technology, the whole *Ellogon* platform was embedded as an ActiveX component under MS Word. However, this embeddance reduced the robustness of the whole integration as sporadic crashes were observed when the user exited Word, as Word failed in some cases to successfully terminate the *Ellogon* ActiveX component. The communication services offered by DDE did not cause any robustness problems, as *Ellogon* runs as a DDE server in a separate process than MS Word. DDE messages are used for bi-directional communication between *Ellogon* and Word as well as for terminating *Ellogon* when the user exits Word.

---

[7] See also at: http://www.iit.demokritos.gr/skel/mitos/.

[8] See also at: http://www.iit.demokritos.gr/skel/en/Projects/SCHEMATOPOIESIS.htm.

### 3.5.4. CROSSMARC

The "CROSS-lingual Multi Agent Retail Comparison" (CROSSMARC[9]) project develops commercial-strength technology for e-retail product comparison. CROSSMARC employs language technology methods for information extraction, which can process pages written in several languages, and can be adapted semi-automatically to new product types.

*Ellogon* is employed by the SKEL laboratory in CROSSMARC, as it provides significant advantages regarding handling of HTML corpora, which is the primary type of corpora for the CROSSMARC project. These advantages include:

- Complete HTML support: *Ellogon* provides facilities for retrieving HTML pages and the contained images from HTTP servers, converting them from any possible character set into Unicode and storing them as well as their images as *Ellogon* documents. Additionally, *Ellogon* provides a complete HTML viewer that can be used to preview the HTML pages.
- HTML aware annotation tools. These tools can be used to annotate HTML document while the annotator sees the HTML preview of the document and not the actual HTML source. An example of such a tool can be seen in Figure 6.
- Perhaps the most significant advantage is the *Ellogon* API: Requiring only an HTML aware tokenisation component, able to identify and separate HTML tokens from non-HTML ones, all subsequent linguistic processing components can be applied to the HTML source *without any modification*, as only tokens corresponding to textual data can be transparently selected through the query facilities of the *Ellogon* API. An example is presented in Figure 7, where a part-of-speech tagger has been applied in an HTML document but only non-HTML tokens have been assigned a part-of-speech category.

## 4. Future Plans

We are continuously working to improve *Ellogon* along many directions. Although *Ellogon* is already highly optimised, we still try to further reduce the memory requirements. Currently, we try to enhance CDM with the ability to selectively load only the needed information from a document in memory instead of the whole document. We are also working towards improving the user interface by adding new features and improving existing ones. At the same time, we are trying to further ease the component development/compilation process. Our goal is to automatically provide some standard compilation tools for the most common platforms, like configure scripts for Unix or Visual C++ project files for Windows, as well the ability to compile components developed in C++ from inside *Ellogon* in a cross-platform way, if a suitable C++ compiler is available.

Future versions of *Ellogon* will provide more ready to use tools for a larger set of common tasks as well as more input filters, like the ability to open Microsoft Word or PDF documents. In addition, we plan to provide an object-oriented version of the current CDM API.

### *Acknowledgements*

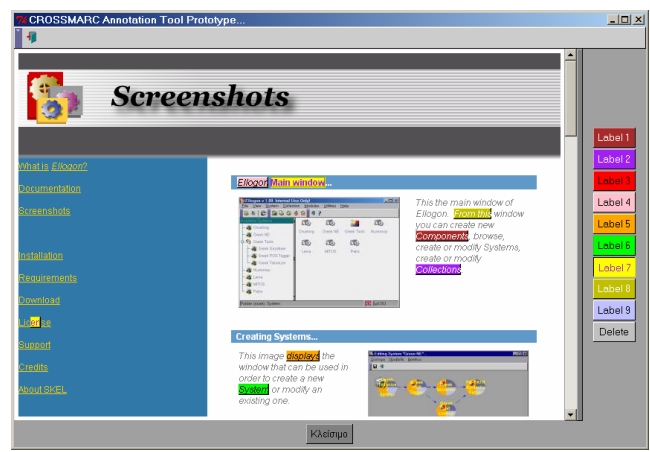

Figure 6: An HTML aware *Ellogon* annotation tool, annotating an *Ellogon* document containing an HTML page and the corresponding images.

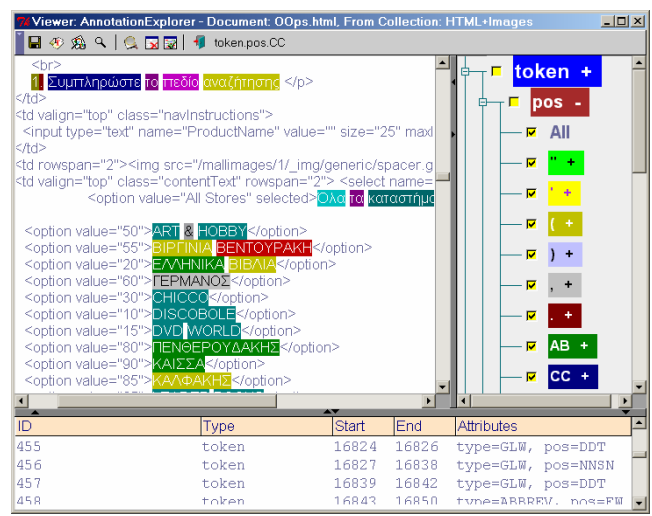

Figure 7: An *Ellogon* viewer presenting the output of a part-of-speech tagger applied on an HTML document.

---

[9] See also at: http://www.iit.demokritos.gr/skel/crossmarc/.

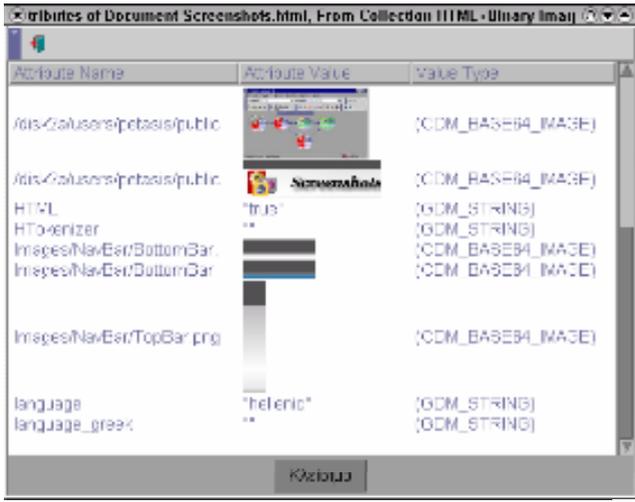

Figure 8: An *Ellogon* viewer presenting attributes contained in a document. Note that the values of some attributes are images.

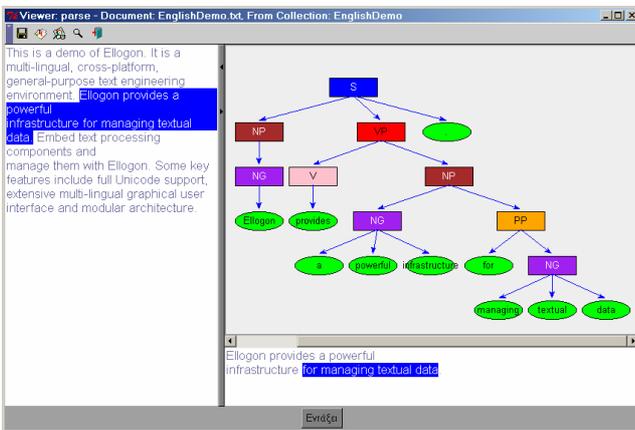

Figure 9: An *Ellogon* viewer showing a parse tree of a simple sentence.